\documentclass{article}

  \usepackage[preprint]{neurips_2026}

\usepackage[utf8]{inputenc} 
\usepackage[T1]{fontenc}    
\usepackage{hyperref}       
\usepackage{url}            
\usepackage{booktabs}       
\usepackage{amsfonts}       
\usepackage{amsmath}
\usepackage{amssymb}
\usepackage{mathtools}
\usepackage{amsthm}
\usepackage{nicefrac}       
\usepackage{microtype}      
\usepackage{xcolor}         
\usepackage{graphicx}
\usepackage{subcaption}
\usepackage{float}
\usepackage{multirow}
\usepackage{enumitem}

\usepackage[capitalize,noabbrev]{cleveref}

\theoremstyle{plain}

\theoremstyle{definition}

\theoremstyle{remark}

\DeclareRobustCommand{\simgym}{{\fontfamily{cmr}\selectfont\textsc{SimGym}}}

\title{SimGym: A Framework for A/B Test Simulation in E-Commerce with Traffic-Grounded VLM Agents}

\author{
Han Li, Vibhor Malik, Zahra Zanjani Foumani, Alberto Castelo\thanks{Corresponding Author:
alberto.castelo@shopify.com}, Shuang Xie, Ailin Fan,\\
\textbf{Keat Yang Koay, Yuanzheng Zhu, Meysam Feghhi, Ronie Uliana, Zhaoyu Zhang,}\\
\textbf{Angelo Ocana Martins, Mingyu Zhao, Francis Pelland, Jonathan Faerman, Nikolas LeBlanc,}\\
\textbf{Aaron Glazer, Andrew McNamara, Zhong Wu, Lingyun Wang}\thanks{Corresponding Author:lingyun.wang@shopify.com} \\ Shopify\\
  Bellevue, Washington, USA \\}

\begin{document}

\maketitle

\begin{abstract}
A/B testing remains the gold standard for evaluating modifications to e-commerce storefronts, yet it diverts traffic, requires weeks to reach statistical significance, and risks degrading user experience. We present \simgym, a framework for simulating A/B tests on e-commerce storefronts using vision-language model (VLM) agents operating in a live browser. The framework comprises three key components: (a) a traffic-grounded persona generation pipeline that derives per-shop buyer archetypes and intents from production clickstream data; (b) a live-browser agent architecture that combines multimodal perception over visual and browser-structured observations with episodic memory and guardrails to conduct coherent shopping sessions across control and treatment storefronts; and (c) an evaluation protocol that compares simulated outcome shifts with observed shifts in real buyer behavior. We validate \simgym~on A/B tests of visually driven UI theme changes from a major e-commerce platform across diverse storefronts and product categories. Empirical results show that \simgym~agents achieve strong agreement with observed outcome shifts, attaining 77\% directional alignment with add-to-cart shifts observed across interface variants in real-buyer traffic. It reduces experimental cycles from weeks to under an hour, enabling rapid experimentation without exposing real buyers to candidate variants.
\end{abstract}

\section{Introduction} \label{sec: intro}
A/B testing remains the gold standard for evaluating modifications in e-commerce storefronts, including UI and theme changes, while enabling data-driven decisions that directly affect conversion rates and revenue. However, traditional A/B testing often incurs substantial costs. Diverting traffic to experimental variants exposes real users to potentially suboptimal experiences, achieving statistical significance often requires weeks of data collection, and unsuccessful treatments can degrade customer experience before they are detected and rolled back. These limitations make merchants more risk-averse in practice: instead of testing bold redesigns with potentially large effects on conversion, merchants often favor incremental, low-risk changes, which constrain innovation and hinder business growth.
Solving the aforementioned challenges motivates a natural question: can \textbf{\textit{synthetic buyers}} be used to pre-test interface modifications before those variants are exposed to real customers? Recent advances in large language model (LLM)-based and vision-language model (VLM)-based agents make this possibility increasingly realistic. Recent work has shown that such agents can navigate complex online interfaces and execute multi-step tasks across diverse browser environments \cite{zhou2023webarena,deng2023mind2web,chezelles2024browsergym,pan2024webcanvas}. In parallel, persona and profile-conditioned agents have begun to exhibit more realistic patterns of user behavior \cite{park2023generative,wang2023recagent,zhang2024generative}. VLMs further strengthen this trend by enabling agents to reason over rendered webpages rather than relying solely on textual page structure, thereby narrowing the gap between agent perception and the visual experience of real shoppers. In e-commerce and design-evaluation settings, systems such as PAARS \cite{mansour2025paars}, Shop-R1 \cite{zhang2025shop}, and Customer-R1 \cite{wang2025customerr1} suggest that agent behavior can be aligned with historical customer data, while AgentA/B \cite{wang2025agenta}, UXAgent \cite{lu2025uxagent}, and SimAB \cite{rieder2026simab} take further steps toward synthetic experimentation and usability testing using persona-conditioned LLM agents.

Despite these advances, existing work remains limited in two important respects. First, prior research has yet to establish an end-to-end framework for synthetic A/B testing in real e-commerce settings. Such a framework must comprise three key components: (a) a persona generation pipeline that grounds synthetic buyers in observed traffic and persona distributions, ensuring that the simulated population reflects the heterogeneity of real users; (b) a stable browser-based interaction environment in which agent behavior can be evaluated under the same live interface conditions encountered by actual customers; and (c) a reliable evaluation module for validating simulated outcomes under interface interventions. Although prior studies have addressed these components only in isolation \cite{mansour2025paars,zhou2023webarena,chezelles2024browsergym,wang2025agenta,lu2025uxagent,rieder2026simab,wang2025customerr1}, they have not integrated them into a unified framework. Consequently, the field still lacks a rigorous basis for conducting synthetic A/B testing on real e-commerce use cases.

Second, the practical utility of the end-to-end simulation workflow depends on how well synthetic agents can predict real user responses to interface changes. However, prior work has not demonstrated such predictive validity in heterogeneous real e-commerce settings. Specifically, existing approaches primarily focus on standardized offline benchmarks such as WebShop \cite{yao2022webshop} and ShoppingBench \cite{wang2025shoppingbench}, behavioral-similarity analyses on clickstream traces \cite{sun2025llm,wang2025customerr1}, static-screenshot A/B simulation without live-browser interaction or magnitude validation \cite{rieder2026simab}, or live-browser usability simulation without validation against measured human intervention effects \cite{lu2025uxagent}. Consequently, none of these approaches thoroughly evaluate whether simulated outcomes align with the observed shifts under real per-shop A/B interventions, leaving a critical gap: an agent could behave reasonably at the level of navigation yet still fail to predict actual conversion shifts, thereby offering limited value for pre-testing decisions in real e-commerce applications.


\begin{figure*}[t]
  \begin{center}
    \centerline{\includegraphics[width=0.8\textwidth]{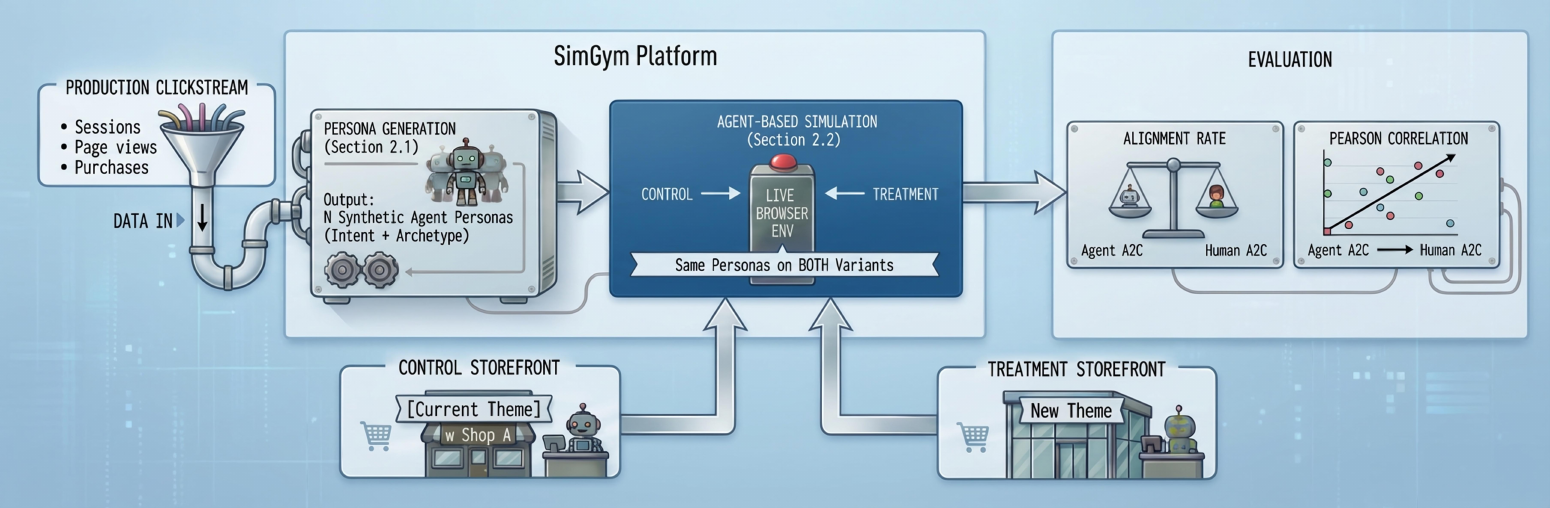}}
    \caption{\simgym~framework overview.}
    \label{fig:proposed_framework}
  \end{center}
  \vskip -0.5in
\end{figure*}

In this paper, we present \simgym, a framework for A/B test simulation in which VLM-powered browser agents, grounded in a merchant's observed customer distribution, autonomously browse two variants of the same online storefront. Unlike prior offline simulation approaches that operate on static datasets \cite{wang2025shoppingbench,zhang2025shop}, \simgym~agents interact with live storefronts in a full browser environment, requiring real-time multimodal perception, memory management, and action execution. By aggregating agent behaviors and decision traces, \simgym~produces comparative metrics that quantify how interface changes, such as visually driven theme changes, affect add-to-cart (A2C) performance. \Cref{fig:proposed_framework} illustrates the end-to-end framework. We summarize our contributions as follows:
\begin{itemize}[leftmargin=*,nosep]
    \item We present an end-to-end framework, \simgym, for synthetic A/B test simulation and evaluation with VLM-powered browser agents. The framework is fully modular, with components that can be independently customized, replaced, or extended.    
    \item We introduce a novel technical realization of \simgym, combining a traffic-grounded synthetic-buyer construction pipeline with a multimodal live-browser agent architecture. The pipeline transforms production clickstreams into shop-specific buyer archetypes and shopping intents, while the agent architecture integrates visual observations, textual webpage representations, episodic memory, and execution guardrails to support coherent, persona-grounded shopping behavior across diverse real-world storefronts.
    \item We validate the framework on real-world, visually driven theme-change use cases from a production e-commerce platform by comparing simulated predictions against observed human behavioral outcomes. Empirical evaluation demonstrates strong directional alignment and correlation between simulated and observed outcome shifts across diverse storefronts and product categories.
\end{itemize}
\section{SimGym Framework} \label{sec: methods}
As shown in \Cref{fig:proposed_framework}, the \simgym~framework consists of three key components: (a) a traffic-grounded persona generation pipeline (\Cref{sec: intent_persona_pipeline}); (b) a multimodal live-browser agent architecture (\Cref{sec: agent_arch}); and (c) an evaluation protocol (\Cref{sec: eval_protocol_overview}). \simgym~ is fully modular, and each component can be independently customized or extended.

\subsection{Persona Generation} \label{sec: intent_persona_pipeline}
The persona generation pipeline derives a population of synthetic shoppers from merchant-specific clickstream traces through a six-stage process. Each synthetic shopper is represented by a \textbf{buyer persona}, which we define as the pairing of a \textbf{buyer archetype} (a multi-dimensional behavioral and values profile) with a \textbf{buyer intent} (a sampled product target paired with a purchase-decision guide). Rather than relying on hand-crafted personas or generic shopper archetypes, the pipeline infers shopper heterogeneity from observed session and buyer-level behavior. At a high level, this process comprises: (a) clustering sessions to identify recurring purchase patterns; (b) extracting product preferences and generating buyer intents; (c) aggregating buyer-level behavioral signals to construct buyer archetypes; and (d) composing intent and archetype into the buyer persona that drives the live-browser agent.

This design enables \simgym~to preserve each storefront's empirical customer distribution, align agent profiles with observed engagement and conversion patterns, and scale persona construction across diverse merchants without manual tuning. We describe the implementation details in \Cref{sec: persona-pipe}.

\subsection{Agent Architecture} \label{sec: agent_arch}
Unlike existing offline simulation approaches \cite{zhang2025shop}, \simgym~adopts a VLM-powered browser agent architecture that executes shopping sessions directly on live storefronts through a full browser environment. To operate reliably across heterogeneous and dynamically changing e-commerce interfaces, this architecture combines multimodal web perception, action planning, memory management, and guarded browser execution, as shown in \Cref{fig:agent_arch}. We describe each component below.

\textbf{Multimodal Web Perception.} \simgym~simulation agents observe each page through two complementary modalities provided to the VLM at every decision step: (a) textual web structure information, represented by a DOM-derived accessibility tree that abstracts the page into a hierarchy of elements, such as headings, buttons, and links, with unique reference IDs mapped to DOM locations for unambiguous action targeting \cite{deng2023mind2web,zhou2023webarena}; and (b) visual page perception, represented by a page screenshot that captures the rendered storefront experience as it would be presented to a real shopper, including layout, imagery, typography, color, and visual emphasis.

The accessibility tree provides a compact, action-oriented representation that preserves the semantic structure needed for navigation and element selection, while the screenshot complements this representation with the rendered visual context conveyed by storefront design, promotional content, and product imagery. Combining these modalities allows \simgym~simulation agents to reason about interface changes whose effects are primarily visual, such as hero redesigns, banner prominence, and color-led merchandising, while maintaining stable, DOM-grounded action execution across heterogeneous storefront designs.


\textbf{Action Planning.} \simgym~simulation agents follow an observe-plan-act loop commonly used in autonomous agent architectures \cite{park2023generative}. In this design, the simulation agent serves as a high-level planner: at each decision step, it determines whether the shopping session should terminate and, if not, generates the next browser-level action to be executed by a dedicated browser controller. To support goal-directed and persona-conditioned decisions, \simgym~assembles a state-conditioned planning context for the VLM that contains: (a) the assigned shopping goal; (b) the generated buyer persona (see \Cref{sec: persona-pipe}), including exploration depth, price sensitivity, product preferences, and value orientation; (c) session memory summarizing prior actions, observations, and intermediate outcomes; (d) the current browser state, represented by the URL, the DOM-derived accessibility tree, and the page screenshot; and (e) guardrails specifying permissible checkout and cart operations.

Given this context, each decision step proceeds through three stages: (a) \emph{observe}, where the agent observes the current page state and retrieves the accumulated session memory; (b) \emph{plan}, where the VLM produces a schema-constrained response containing its reasoning, a termination decision, and a proposed next action if the session should continue; and (c) \emph{act}, where the system either terminates the session or passes the predicted action to the dedicated browser controller for execution on the live page. The action specification includes the action type, target element, and required arguments, making the model output directly parseable by downstream execution logic.

After execution, the controller records the resulting page transition, action outcome, and relevant observations, then appends this information to session memory for subsequent planning steps. This loop repeats until the simulation agent completes its shopping goal or triggers a guardrail. The process keeps each decision grounded in the current page state while preserving session-level continuity across the full browsing trajectory. We provide an example agent trajectory in Appendix \ref{sec: agent_trace}.

\textbf{Memory Management.} The simulation agents maintain an episodic session memory that accumulates the complete browsing trajectory, including the initial navigation context, prior observations, model reasoning and decisions, executed actions, action outcomes, and any error states. This memory is incorporated into the planning context at each decision step, allowing the agent to maintain temporal coherence across multi-step shopping sessions. By conditioning on previously viewed products, attempted interactions, and failed actions, the agent can reduce redundant behavior and make more consistent decisions across the browsing trajectory, mirroring how real shoppers maintain and use browsing context during product exploration.

\textbf{Guardrails.} To improve reliability in open-ended web environments, \simgym~uses guardrails that constrain execution and support recovery from common failure modes. These mechanisms include: (a) loop protection, which detects repeated identical actions and prevents unbounded execution \cite{pan2024webcanvas}; (b) step and time budgets, which bound the duration of each simulation; (c) model retry logic, which retries failed VLM calls with error context; and (d) error propagation, which passes browser execution failures back to the agent to support informed recovery rather than blind retry. These guardrails ensure that simulations recover or terminate gracefully when agents encounter unexpected page states, execution failures, or reasoning errors.



\subsection{Evaluation Protocol} \label{sec: eval_protocol_overview}
The evaluation protocol connects simulated shopping behavior to behavioral shifts observed in human traffic on production storefronts, and is designed to evaluate whether simulated agents predict the effect of real interface changes, rather than merely producing plausible browsing traces. The protocol consists of two stages: (a) ground-truth construction that identifies paired control and treatment storefront variants with sufficient traffic and substantive UI differences, while excluding pairs whose A2C signal would be dominated by non-theme factors such as overlapping promotional campaigns, merchandising or assortment shifts, pricing changes, and new-shop ramp-up periods; and (b) predictive-validity measurement that compares simulated and observed human shifts in add-to-cart behavior using alignment rate and Pearson correlation. 

The protocol operates only on per-agent session logs and observed storefront outcomes, and is therefore agnostic to the upstream persona generation pipeline and browser-agent architecture. Alternative persona generators or agent implementations can be integrated and evaluated using the same evaluation protocol under the modularized design. We describe the ground-truth dataset and metric definitions in detail in \Cref{sec: data_and_eval}.


\section{Persona Generation}\label{sec: persona-pipe}
Building synthetic shoppers that can predict real customer behavior requires agent profiles grounded in each merchant's traffic distribution. Storefronts often serve distinct buyer populations with heterogeneous intents, preferences, and browsing behaviors. For example, a boutique fitness brand can exhibit a significantly different traffic composition than a sporting-goods retailer. Unlike traditional persona generation which relies on UX intuition with fictional personas, we directly extract shopper intents and behavioral preferences from each shop's clickstream data. To prevent evaluation leakage, persona construction uses historical clickstreams disjoint from the A/B sessions used for ground-truth A2C estimation. Capturing merchant-specific shopper heterogeneity at scale entails two design objectives: (a) generated behavioral preferences should reflect the empirical distribution of buyer types observed on each storefront; and (b) the pipeline should scale across a large and heterogeneous set of merchants without manual persona engineering or store-specific tuning. Guided by these design objectives, we construct agent personas through a six-stage pipeline.

\textbf{Stage 1: Session-Level Clustering.}  We represent each storefront session as a feature vector spanning engagement (duration, event count), exploration depth (product views, distinct items), search behavior, funnel progression (A2C, checkout, purchase), and economic value (cart value, order value). Features are standardized via z-scoring across sessions, and we apply k-means clustering with k-means++ \cite{arthur2007k} initialization. We use $k=5$ selected by an elbow-style analysis of k-means inertia and cluster-size balance. Each session is assigned to its nearest centroid, with the standardized distance retained as a measure of assignment uncertainty.



\textbf{Stage 2: Product Preference Extraction.} For each (shop, cluster) pair, we construct an input record containing shop metadata (name and industry), the cluster identifier, and an aggregate summary of products browsed or purchased by sessions in the cluster. We use GPT-5 \cite{singh2025openai} to extract product preferences from this record and output a structured JSON object with three fields: (a) \textit{product categories}, listing up to ten broad categories such as ``sneakers'' or ``athletic wear''; (b) \textit{individual products}, listing up to ten frequently browsed or purchased products in the cluster; and (c) a brief reasoning. We restrict categories to generic descriptors rather than shop-specific product names to improve reuse in downstream persona generation.


\textbf{Stage 3: Buyer Intent Generation.} We generate a structured shopping intent for each agent with two components: (a) \emph{product target}, sampled from the Stage 2 product categories associated with the agent's cluster and paired with a buyer archetype from the same cluster, so that agents share a behavioral profile but vary in shopping targets; and (b) \emph{purchase-decision guide}, a fixed instruction that asks the agent to research available options, defer purchase by default, and add a product to cart only when the product clearly fits the assigned archetype and the storefront experience provides sufficient purchase confidence. To keep intents reusable across storefront variants, we exclude bundles, sizes, discounts, and UI-specific details from the product target.



\textbf{Stage 4: Buyer Behavior Aggregation.} For each (shop, cluster) pair, let $n$ denote the number of simulated agents allocated to that cluster. We select the $n$ sessions nearest to the cluster centroid, using centroid proximity to obtain representative behavioral examples for agent construction. We then aggregate the selected sessions at the buyer level to compute cross-session summaries including session counts, funnel outcomes (A2C, checkout, purchase), average cart and order values, and product interaction histories.


\textbf{Stage 5: Buyer Archetype Construction.} We construct buyer archetypes as multi-dimensional profiles rather than single mutually exclusive labels, allowing multiple traits to coexist (e.g., price-sensitive and ethics-oriented). Each buyer is represented along five continuous dimensions grouped into behavioral and value dimensions (\Cref{tab:archetype-dimensions}). Behavioral dimensions describe how the buyer shops and are computed from aggregated clickstream statistics. Value dimensions describe what product attributes appear to drive choices. For each value axis, we identify relevant products using predefined keyword sets (e.g., ``handcrafted'', ``organic'', ``commercial grade'') and compare their prevalence among browsed and purchased items, separating general interest from revealed purchase preference. Finally, we use GPT-5 to output a structured JSON with a score for each archetype dimension and a concise rationale. The full schema and an example output are shown in \Cref{fig:archetype-extraction}.

\begin{table*}[t]
  \caption{Buyer Archetype Dimensions}
  \vskip -0.1in
  \label{tab:archetype-dimensions}
  \begin{center}
    \begin{small}
        \begin{tabular}{llp{10cm}}
          \toprule
          Dimension & Type & Description \\
          \midrule
          Price sensitivity & Behavioral & Premium / mid-range / budget, inferred from browsed vs.\ purchased price gaps with category-aware normalization \\
          Exploration depth & Behavioral & 0--1 score based on session duration, searches, and product views \\
          Premium focus & Values & Attention to luxury, craftsmanship, prestige \\
          Performance focus & Values & Emphasis on durability, reliability, specifications \\
          Ethics focus & Values & Interest in sustainability, ethical sourcing \\
          \bottomrule
        \end{tabular}
    \end{small}
  \end{center}
  \vskip -0.3in
\end{table*}


\textbf{Stage 6: Prompt Composition.} We form each agent persona by pairing a Stage 3 shopping intent with a Stage 5 buyer archetype from the same cluster. The final prompt combines the product target, purchase-decision guide, behavioral and value profiles, and cluster-level product preferences from Stage 2. This yields personas that reflect observed product interests and multi-dimensional buyer traits. \Cref{fig: output_persna} shows an example persona output.

\section{Evaluation Protocol}\label{sec: data_and_eval}
In this section, we describe the evaluation protocol along two dimensions: (a) constructing a ground-truth dataset to capture how human shoppers respond to storefront changes; and (b) defining evaluation metrics to measure predictive validity across diverse buyer types.

\subsection{Ground Truth Construction} \label{sec: data_generation}

To assess the predictive validity of \simgym, we construct a ground-truth dataset from real A/B experiments of theme changes on the e-commerce platform. Each experiment compares a control storefront against a treatment storefront that differs in visual presentation or layout, allowing us to measure human A2C shifts directly from real buyer traffic exposed to the two variants. We select shops using two criteria: (a) sufficient traffic in both control and treatment groups to support reliable human A2C estimation; and (b) substantial visual or layout differences between the paired storefront variants.


Since the paired storefront variants vary in magnitude, from minor styling tweaks to complete layout redesigns, we employ Gemini 3 Pro \cite{googledeepmind2025gemini3pro} as a strong VLM evaluator to jointly inspect page screenshots and the parsed DOM, characterize the change in each control-treatment pair, and stratify the dataset by change magnitude. We additionally apply quality-control filters to exclude pairs whose A2C signal would be dominated by non-theme factors (overlapping promotional campaigns, merchandising or assortment shifts, pricing changes, and new-shop ramp-up periods), and require that the remaining theme changes are non-trivial. This yields a final dataset of $50$ shops, each containing a paired control-treatment storefront comparison, a summary of the theme change, and observed human A2C shifts between variants.
\subsection{Evaluation Methodology} \label{sec: eval}
We evaluate \simgym~by measuring predictive validity against real human outcomes. Rather than comparing exact browsing trajectories, we assess whether simulated agents predict the behavioral shifts induced by storefront changes. We use A2C rate as the primary outcome, as it directly reflects purchase intent and corresponds to a key optimization target for merchants. Our evaluation focuses on two dimensions of predictive validity: (a) for directional validity, we use alignment rate as the percentage of shops for which the sign of the agent-predicted A2C shift between storefront variants matches the sign of the observed human A2C shift; and (b) for magnitude validity, we use the Pearson correlation coefficient between agent-predicted and human-observed A2C shifts across shops. We report 95\% confidence intervals for both metrics over the $n{=}50$ shops. Computation details are in \Cref{appendix:ci}.

\vspace{-0.15in}
\section{Experiments}
\label{sec:experiments-results}

\begin{figure}[t]
  \begin{center}
    \centerline{\includegraphics[width=0.9\columnwidth]{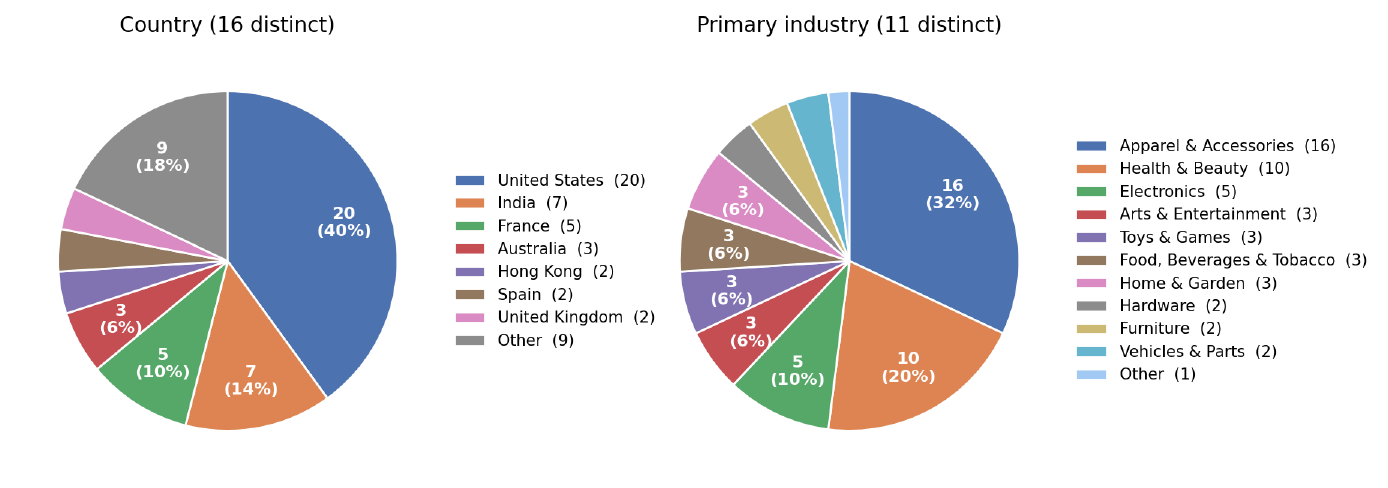}}
    \vskip -0.2in
    \caption{Dataset distribution of the 50-storefront golden set spanning 16 countries and 11 industries. It includes smaller markets and specialized categories to support evaluation across diverse storefronts.}

    \label{fig:dataset}
  \end{center}
  \vskip -0.3in
\end{figure}

In this section, we evaluate \simgym~simulations on 50 real-world shops spanning 16 countries and 11 product categories (\Cref{fig:dataset}). These 50 shops have ground-truth A2C shifts constructed from observed human behavior across control and treatment storefront variants with visually driven theme changes, using the procedure described in \Cref{sec: data_generation}. We report all metrics on the \textit{skimmers} cohort, the largest non-bouncer traffic segment, which accounts for $64.5\%$ of all engaged sessions. This cohort exhibits substantive purchase intent with a roughly $9.5\%$ session-level A2C rate, and its A2C shifts are highly correlated with whole-shop A2C shifts.\footnote{A2C shift denotes the change in human A2C rate between the control and treatment storefront variants. Across the 50 shops, A2C shifts measured on the skimmers cohort are highly correlated with those measured on whole-shop traffic after excluding immediate bouncers ($r \approx 0.89$; Appendix \ref{appendix:cohort-coverage}).}

\begin{table}[t]
  \caption{Overall simulation evaluation under different LLMs for agent decision making.}
  \label{tab:overall_eval}
  \begin{center}
    \begin{small}
    \setlength{\tabcolsep}{4pt}
        \begin{tabular}{lccccc}
          \toprule
          Model                      & Align. (\%)   & Align.\ 95\% CI & Corr.         & Corr.\ 95\% CI  & Runtime/shop \\
          \midrule
          Gemini 3 Flash             & \textbf{77.0} & [66.0, 87.0]    & \textbf{0.55} & [0.32,  0.72]  & 5.3 min      \\
          Gemini 3 Flash (text only) & 70.0          & [58.0, 82.0]    & 0.49          & [0.25,  0.68]  & 4.5 min      \\
          GPT-OSS (text only)        & 59.0          & [47.0, 71.0]    & 0.41          & [0.15,  0.62]  & 12.8 min     \\
          \bottomrule
        \end{tabular}
    \end{small}
  \end{center}
  \vskip -0.1in
\end{table}

Each agent uses the Stagehand browser-automation library \cite{browserbasestagehand} as the browser controller for operating live storefronts. In our experiments, we use proprietary models for agent decision making through public APIs (Gemini 3 Flash \cite{gemini3flash2025}) and serve open-source models on an internal cluster with $40$ NVIDIA B200 GPUs (GPT-OSS \cite{agarwal2025gpt}). In each simulation, we run $600$ agents per shop. To reduce stochastic variation, we repeat each shop-level simulation twice and report metrics averaged over the two trials.



\subsection{Overall Evaluation}
\label{sec:overall-eval}

Table~\ref{tab:overall_eval} reports the overall simulation performance under different models used for agent decision making. We make the following key observations. First, Gemini 3 Flash with visual integration achieves the strongest simulation quality, reaching $77\%$ directional alignment and $0.55$ correlation with observed human A2C shifts. This indicates that a strong backbone model paired with full multimodal agent perception provides the best predictive agreement with real behavioral responses. Second, the comparison between Gemini 3 Flash with and without visual input demonstrates the value of multimodal perception, as the vision-enabled agent improves both alignment ($77\%$ vs. $70\%$) and correlation ($0.55$ vs. $0.49$) over its text-only counterpart. This suggests that visually salient storefront features (e.g., image quality, layout density, and trust cues) provide behavioral signals that are only coarsely captured by DOM structure alone and help agents better predict buyer responses to interface changes. Third, GPT-OSS underperforms the Gemini 3 Flash configurations, suggesting that a performance gap remains between open-source and leading proprietary models for agent decision making. However, the GPT-OSS text-only agent still achieves $59\%$ alignment and $0.41$ correlation, producing meaningful simulation signals despite lacking visual input. This suggests that \simgym~maintains reasonable predictive validity with less capable open-source models, highlighting the robustness of the framework beyond proprietary multimodal systems. Finally, \simgym~supports efficient per-shop simulation across model configurations. With a 600-agent budget, a single-shop simulation completes in 5.3 minutes for Gemini 3 Flash with vision, 4.5 minutes for text-only Gemini 3 Flash, and under 13 minutes for GPT-OSS. These runtimes suggest that \simgym~ supports rapid storefront evaluation across different agent decision-making models.

\subsection{Ablation Studies}
\label{sec:ablation-studies}
This section presents two key ablations based on the Gemini 3 Flash vision-enabled agent to quantify the impact of \simgym's persona generation and memory management components on overall simulation quality.

\subsubsection{Effect of Persona Generation Strategy}
To quantify the effect of persona and intent inputs from the persona generation pipeline (see \Cref{sec: persona-pipe}), we compare three variants on the same 50-shop golden set: \textit{Full Persona} (a buyer archetype with a shopping intent), \textit{Shopping Intent Only} (a target product with a purchase-decision guide), and \textit{Product Only} (only the target product to shop for without buyer archetype or purchase guide).

\begin{figure}[t]
  \begin{center}
    \begin{subfigure}{0.22\columnwidth}
      \centering
      \includegraphics[width=\columnwidth]{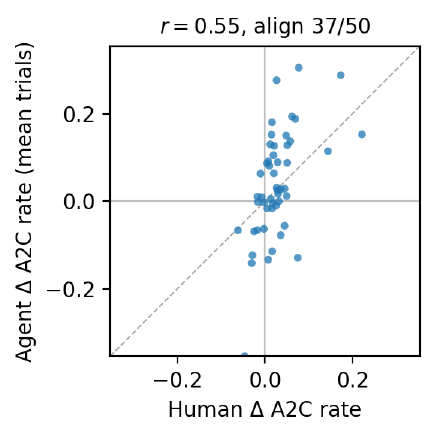}
      \caption{Full Persona}
      \label{fig:pred-custom}
    \end{subfigure}
    \vspace{0.5em}
    \begin{subfigure}{0.22\columnwidth}
      \centering
      \includegraphics[width=\columnwidth]{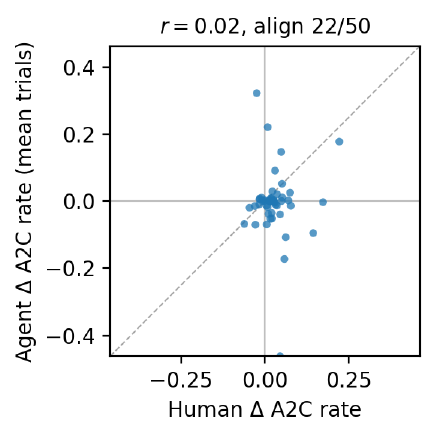}
      \caption{Shopping Intent Only}
      \label{fig:pred-intent-only}
    \end{subfigure}
    \begin{subfigure}{0.22\columnwidth}
      \centering
      \includegraphics[width=\columnwidth]{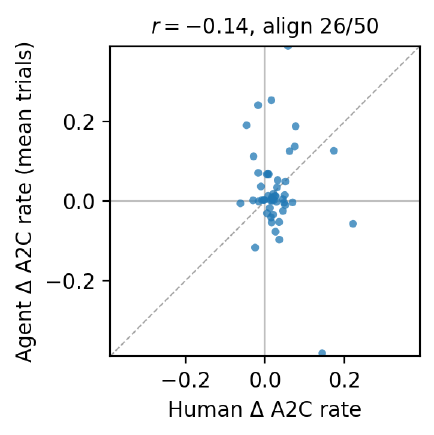}
      \caption{Product Only}
      \label{fig:pred-product-only}
    \end{subfigure}
    \caption{Human-agent agreement in A2C shifts. Each panel plots human-observed versus agent-predicted $\Delta$A2C rates under a persona input variant. The dashed line denotes $y{=}x$.}
    \label{fig:four-panel}
  \end{center}
  \vskip -0.3in
\end{figure}

\Cref{tab:persona} demonstrates that the full persona representation is central to predictive validity. First, \textit{Full Persona} achieves the strongest performance, with $77\%$ directional alignment and $0.55$ correlation with observed human A2C shifts. Second, removing the buyer archetype substantially degrades both metrics: \textit{Shopping Intent Only} drops to $44\%$ alignment with near-zero correlation, indicating that product targets and purchase intentions alone do not capture sufficient behavioral heterogeneity. Finally, \textit{Product Only} performs worst, with chance-level alignment ($51\%$) and the lowest correlation. These results indicate that accurate simulation depends on population-specific behavioral variation encoded by the buyer archetype. \Cref{fig:four-panel} further supports this pattern at the shop level. \textit{Full Persona} shows the strongest human-agent agreement, while \textit{Shopping Intent Only} and \textit{Product Only} scatter diffusely with weak correspondence to human A2C shifts.


\begin{table}[t]
  \caption{Effect of persona and intent inputs on predictive validity.}
  \label{tab:persona}
  \begin{center}
    \begin{small}
        \begin{tabular}{lcccc}
          \toprule
          Configuration       & Alignment (\%) & Align.\ 95\% CI & Correlation   & Corr.\ 95\% CI \\
          \midrule
          Full Persona        & \textbf{77.0}  & [66.0, 87.0]    & \textbf{0.55} & [0.32,  0.72] \\
          Shopping Intent Only& 44.0           & [32.0, 56.0]    & 0.02          & [-0.26,  0.30] \\
          Product Only        & 51.0           & [38.0, 64.0]    & -0.14         & [-0.41,  0.14] \\
          \bottomrule
        \end{tabular}
    \end{small}
  \end{center}
  \vskip -0.1in
\end{table}

\subsubsection{Effect of Memory Management Strategy}
We study the effect of episodic session memory on simulation quality by comparing the full agent with a variant that removes memory access. In the no-memory setting, the agent receives only the current page state and persona at each step, without access to prior actions or observations.

\Cref{tab:memory_ablation} shows that removing episodic session memory substantially degrades predictive validity. Without memory, directional alignment drops to $42\%$, and correlation collapses to $0$, indicating that the agent no longer recovers either the direction or magnitude of human A2C shifts. This suggests that session memory is essential for coherent multi-step shopping behavior and for producing simulation outcomes that remain predictive of real buyer responses. A failure-mode analysis further attributes this degradation to navigation incoherence. Without memory, $70.7\%$ of diverged agents become stuck in navigation loops. This suggests that memoryless agents often fail to track visited pages or attempted actions, preventing them from reaching the product-evaluation stage where storefront changes can meaningfully affect A2C decisions.


\begin{table}[t]
  \caption{Effect of session memory on predictive validity.}
  \label{tab:memory_ablation}
  \begin{center}
    \begin{small}
        \begin{tabular}{lcccc}
          \toprule
          Configuration & Alignment (\%) & Align.\ 95\% CI & Correlation   & Corr.\ 95\% CI \\
          \midrule
          With Memory   & \textbf{77.0}  & [66.0, 87.0]    & \textbf{0.55} & [0.32,  0.72] \\
          No Memory     & 42.0           & [30.0, 54.0]    & 0.00          & [-0.28,  0.28] \\
          \bottomrule
        \end{tabular}
    \end{small}
  \end{center}
  \vskip -0.1in
\end{table}

\subsection{Sensitivity Analysis}
\label{sec:sample-size}
\begin{figure}[t]
  \begin{center}
  \begin{subfigure}[b]{0.3\columnwidth}
      \centering
      \includegraphics[width=\textwidth]{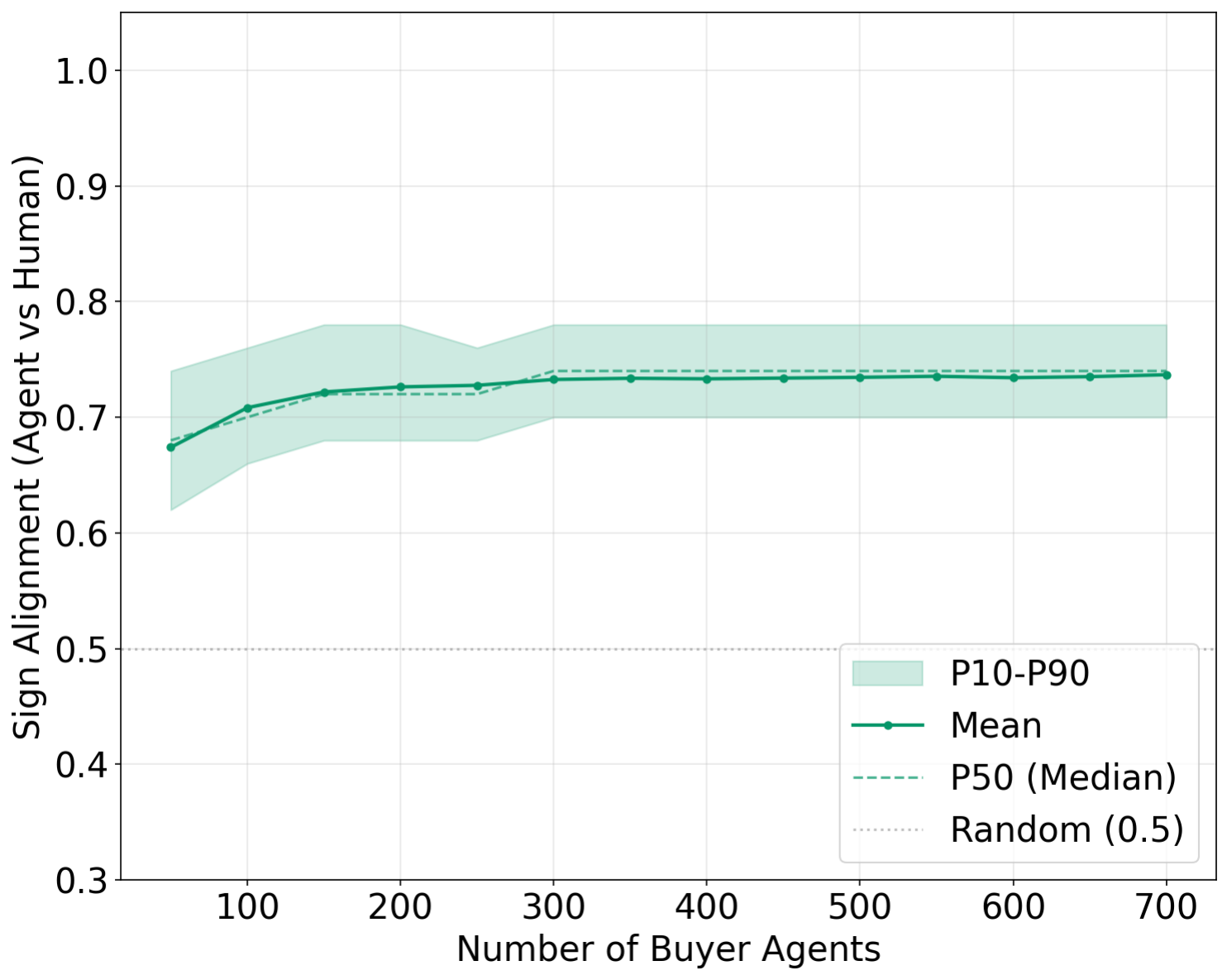}
      \caption{Directional alignment}
      \label{fig:bootstrap_alignment}
    \end{subfigure}
    \begin{subfigure}[b]{0.3\columnwidth}
      \centering
      \includegraphics[width=\textwidth]{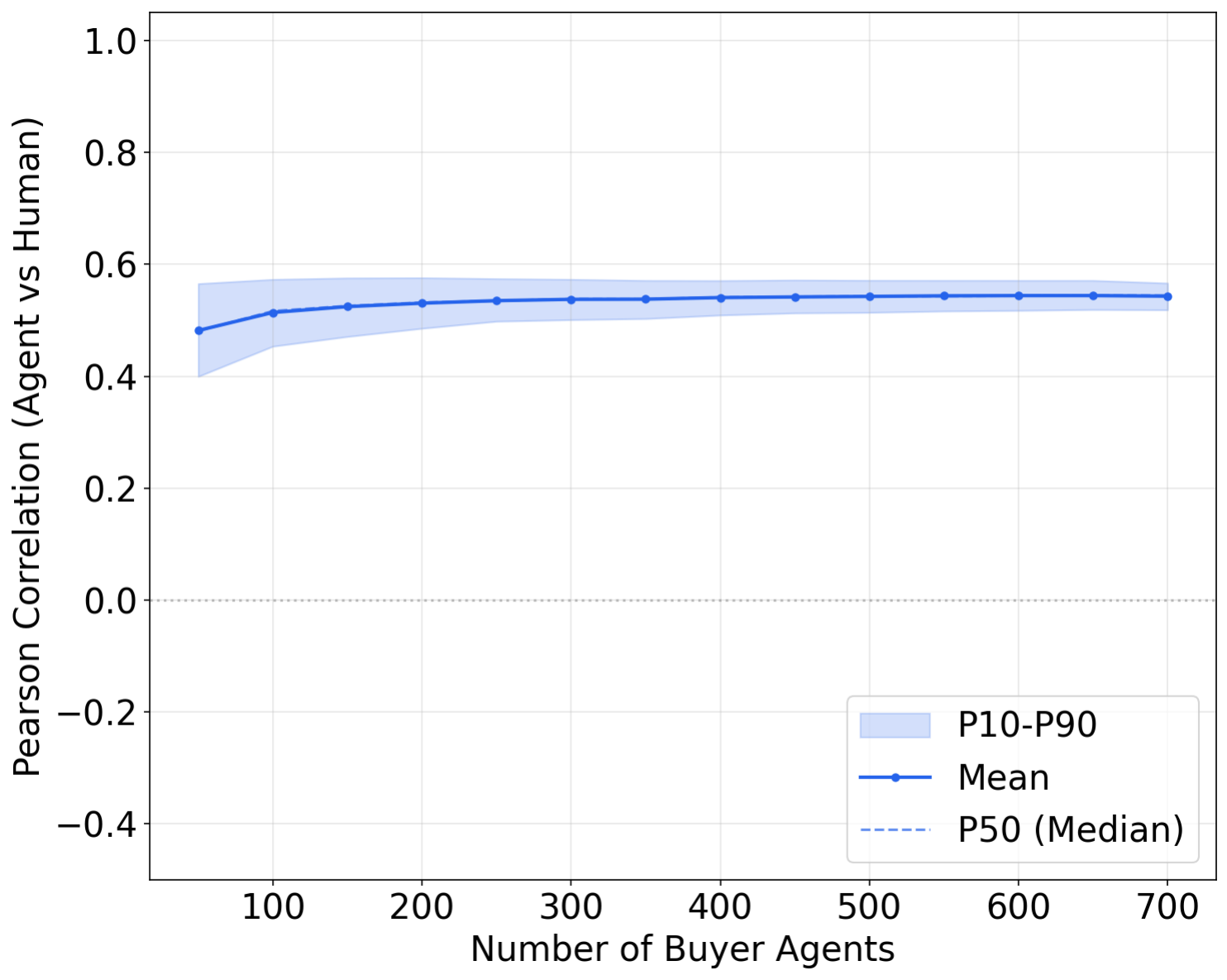}
      \caption{Pearson correlation}
      \label{fig:bootstrap_correlation}
    \end{subfigure}
    \caption{Agent sample-size sensitivity on the $50$-shop golden set. Shaded bands denote the 10th--90th percentile range over $1000$ bootstrap resamples.}    \label{fig:bootstrap}
  \end{center}
  \vskip -0.2in
\end{figure}


Since shop-level simulated A2C shifts are estimated from a finite set of buyer-agent sessions, the agent budget (i.e., the number of agents per shop simulation) determines both estimate variance and simulation cost. We therefore select the agent budget by identifying where additional agents yield diminishing gains in aggregate predictive validity. Using a completed Gemini 3 Flash run with vision and full persona on the $50$-shop golden set, we evaluate finite-sample stability via bootstrap resampling. For each agent budget from $50$ to $700$, we resample buyer-agent sessions with replacement within each shop, recompute shop-level agent $\Delta$A2C, and then recompute cross-shop Pearson correlation and directional alignment against human $\Delta$A2C. We perform $1000$ bootstrap resamples for each agent budget.

\Cref{fig:bootstrap} shows the finite-sample behavior of the evaluation metrics with two key findings: (a) For directional validity, alignment rises from $67\%$ at $50$ agents to $73\%$ by $300$ agents, with little additional improvement through $700$ agents. (b) For magnitude validity, correlation increases from $0.48$ at a budget of $50$ to $0.54$ by $300$, after which gains plateau; the 10th--90th percentile band also narrows from $0.17$ at $50$ agents to $0.05$ at $600$ agents. Based on these findings, we use an agent budget of $600$ in all experiments, which places the simulation beyond the observed stability plateau and provides reliable estimates with additional buffer for finite-sample variability and session-level failures.
\vspace{-0.1in}
\section{Conclusion} \label{sec: conclusion}
We presented \simgym, a framework for e-commerce A/B test simulation that uses traffic-grounded VLM-powered browser agents to interact with live storefront variants and predict the effects of interface changes on user behavior. The framework includes a traffic-grounded persona generation pipeline, a multimodal browser agent architecture with memory and guardrails, and an evaluation protocol that validates simulated outcome shifts against observed human outcomes. Empirical results on 50 real storefront changes show that \simgym~achieves strong agreement with observed human add-to-cart shifts while significantly reducing evaluation cycles, enabling rapid experimentation without exposing real buyers to candidate variants.


\textbf{Limitations and Future Work.}  Our study is conducted on a single major e-commerce platform. While the evaluation covers a 50-shop golden set of curated A/B-tested storefront changes spanning 16 countries and 11 product industries, generalization beyond this setting remains to be tested. We also report results without post-training alignment on human traces, whose impact on predictive validity remains to be validated. Important future directions include extending the framework beyond a single platform and beyond add-to-cart shifts to broader behavioral targets, and improving agent fidelity through post-training alignment on human traces. More broadly, \simgym\ can serve as an inner loop for automated UI exploration, enabling candidate interface changes to be proposed, screened, and prioritized in simulation before merchant-facing experiments.

\nocite{*}
{
\small
\bibliographystyle{plainnat}
\bibliography{ref}
}

\newpage
\appendix
\appendix
\section{Broader Impact}
\simgym~studies offline simulation for A/B testing e-commerce storefronts. Its primary positive impact is to reduce the need to expose real customers to untested interface variants. By allowing merchants to pre-screen potentially risky redesigns in simulation, \simgym~can help reduce user exposure to suboptimal or frustrating shopping experiences, accelerate iteration from multi-week online experiments to substantially faster offline evaluation, and prioritize which variants should be evaluated with real traffic.

We also recognize several potential risks. First, live-browser agents should be clearly separated from real user traffic so that simulated sessions do not contaminate merchant analytics, marketing attribution, or downstream recommender training. Second, automated UI optimization could be misused to reinforce manipulative or deceptive design patterns that improve short-term conversion at the expense of user well-being. Any deployment should therefore couple predictive-validity objectives with explicit ethical, usability, and user-experience constraints. Third, persona construction from production clickstream data is privacy-sensitive. Such data should be processed only under appropriate data-use agreements, with de-identification, aggregation where applicable, minimum-cohort thresholds, and strict access controls to prevent leakage of individual browsing behavior.




\section{Buyer Archetype Construction}
\label{appendix:archetype}
\begin{figure}[H]
  \begin{center}
    \centerline{\includegraphics[width=0.35\columnwidth]{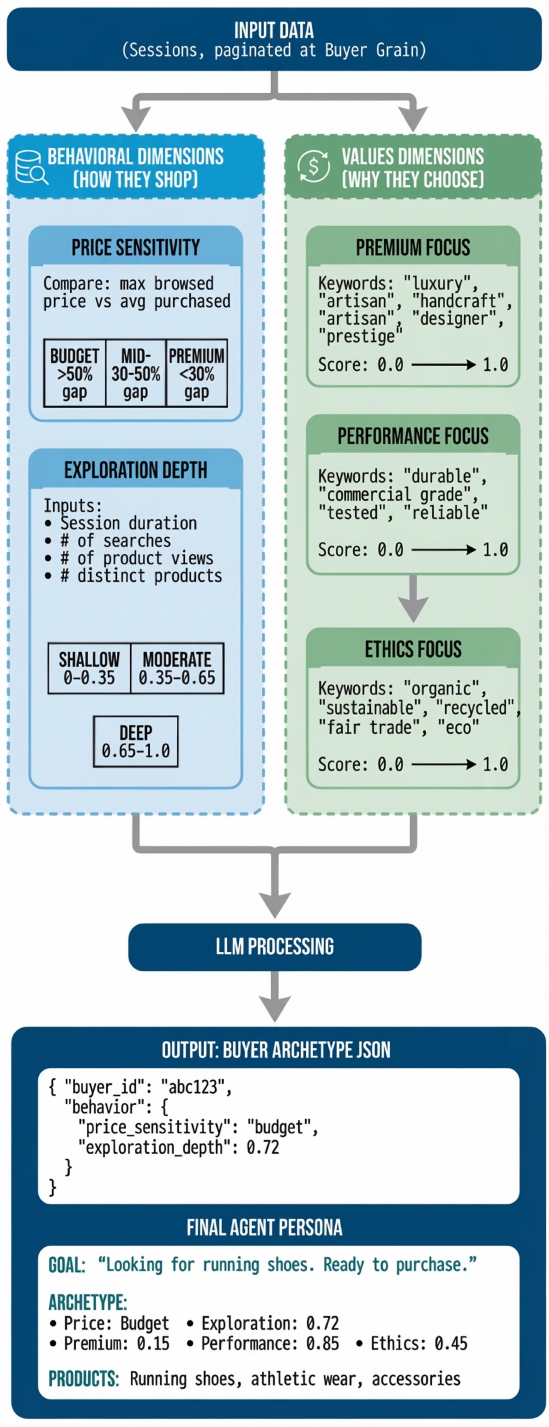}}
    \caption{Buyer Archetype Construction Framework.}
    \label{fig:archetype-extraction}
  \end{center}
  \vskip -0.1in
\end{figure}
\Cref{fig:archetype-extraction} details the buyer archetype construction process. Given aggregated buyer data from 
Stage 4, we compute scores along five continuous dimensions organized into two 
categories:

\textbf{Behavioral dimensions} capture \textit{how} buyers shop:
\begin{itemize}[nosep]
    \item \textit{Price sensitivity}: Measures the gap between maximum browsed 
    price and average purchased price. Buyers are labeled as budget ($>$50\% gap), 
    mid-range (30--50\% gap), or premium ($<$30\% gap), with category-aware 
    normalization to account for product-specific price distributions.
    \item \textit{Exploration depth}: A 0--1 score derived from session duration, 
    search count, and product views, mapped to shallow (0--0.35), moderate 
    (0.35--0.65), or deep (0.65--1.0) exploration regimes.
\end{itemize}

\textbf{Values dimensions} capture \textit{why} buyers choose products:
\begin{itemize}[nosep]
    \item \textit{Premium focus}: Attention to luxury, craftsmanship, and prestige.
    \item \textit{Performance focus}: Emphasis on durability, reliability, and specifications.
    \item \textit{Ethics focus}: Interest in sustainability and ethical sourcing.
\end{itemize}

For each values dimension, we identify products containing relevant keywords 
(e.g., ``handcrafted,'' ``commercial grade,'' ``organic'') and compare the 
proportion in browsed vs.\ purchased items to infer revealed preferences. 
An LLM processes these signals with category-aware taxonomy and deterministic 
decision rules, outputting a structured JSON with scores, confidence estimates, 
and reasoning traces.

\section{Persona Extracted Output}
\label{appendix:intent-output}
\begin{figure}[t]
  \vskip 0.1in
  \begin{center}
        \centering
        \includegraphics[width=0.5\linewidth]{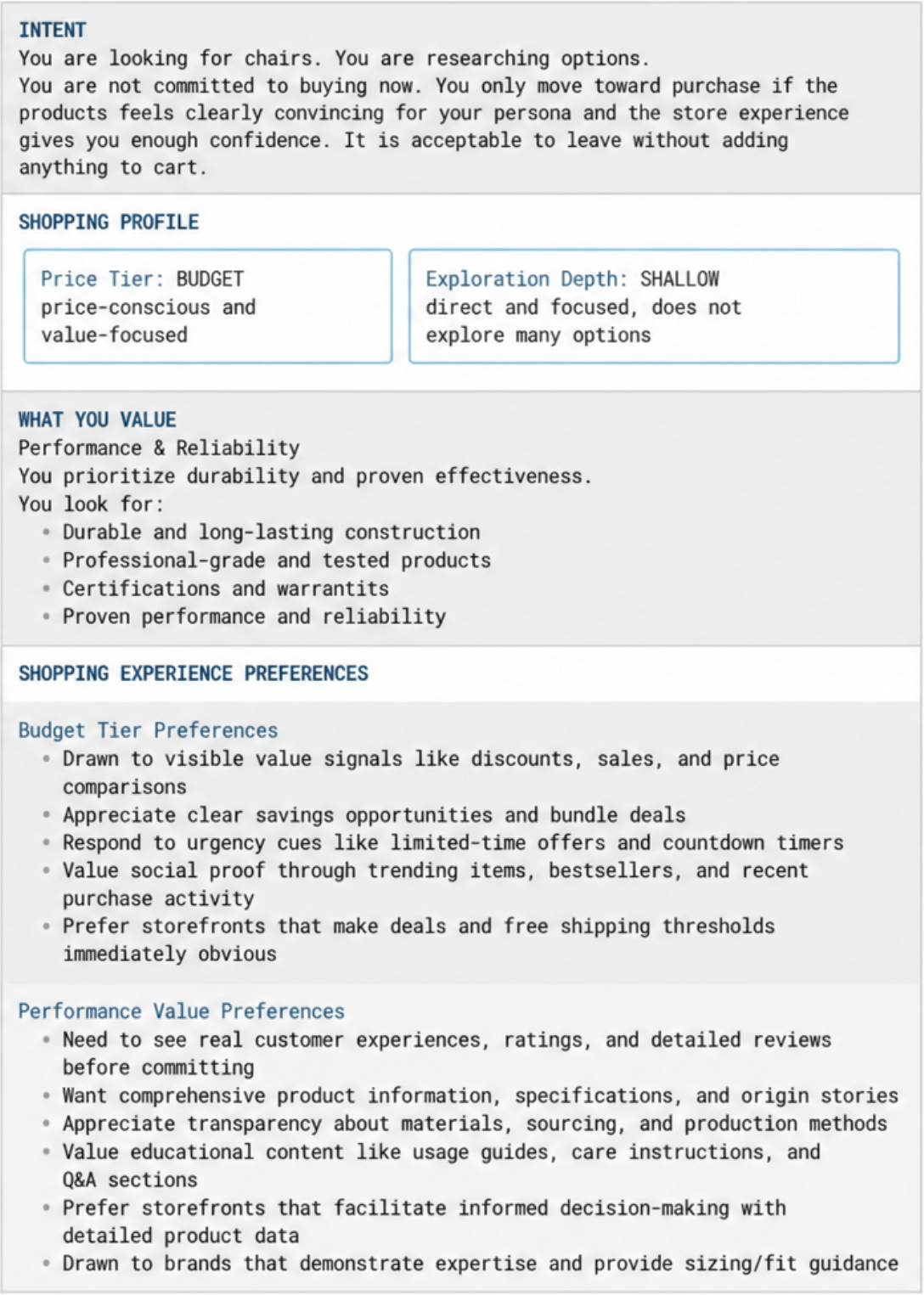}
        \caption{Persona Extracted Output.}
    \label{fig: output_persna}
  \end{center}
  \vskip -0.1in
\end{figure}

\Cref{fig: output_persna} presents a representative output from our persona generation pipeline described in \Cref{sec: intent_persona_pipeline}. The visualization illustrates how the six-stage pipeline transforms raw clickstream data into a structured agent prompt.

The Intent component, generated in Stage 3, pairs a sampled product target (here, ``chairs'') with the fixed shopping guide described in \Cref{sec: persona-pipe}, which anchors purchase decisions to the archetype.
The Shopping Profile encodes two behavioral dimensions derived from Stage 4's buyer behavior aggregation and Stage 5's archetype construction: $(1)$ Price Tier is set to "Budget" (price-conscious and value-focused), reflecting category-aware price sensitivity computed from the gap between browsed and purchased price points; $(2)$ Exploration Depth is classified as "Shallow" (direct and focused), derived from the buyer's session duration, search frequency, and product view counts.

The Values section captures the buyer's "Performance \& Reliability" orientation, one of three values dimensions in our framework (see \Cref{tab:archetype-dimensions}). This classification emerges from Stage 5's analysis of keyword-matched product interaction patterns (e.g., "durable," "professional-grade," "certified").

Finally, the Shopping Experience Preferences operationalize the persona into actionable behavioral guidance. These preferences are derived from the intersection of behavioral and values dimensions: Budget Tier Preferences specify responsiveness to discount signals, social proof, and urgency cues, while Performance Value Preferences emphasize attention to detailed specifications, customer reviews, and transparency about materials. Together, these components enable the agent to exhibit coherent, persona-consistent behavior throughout the shopping session.

\begin{figure}[t]
    \centering
    \includegraphics[width=0.65\linewidth]{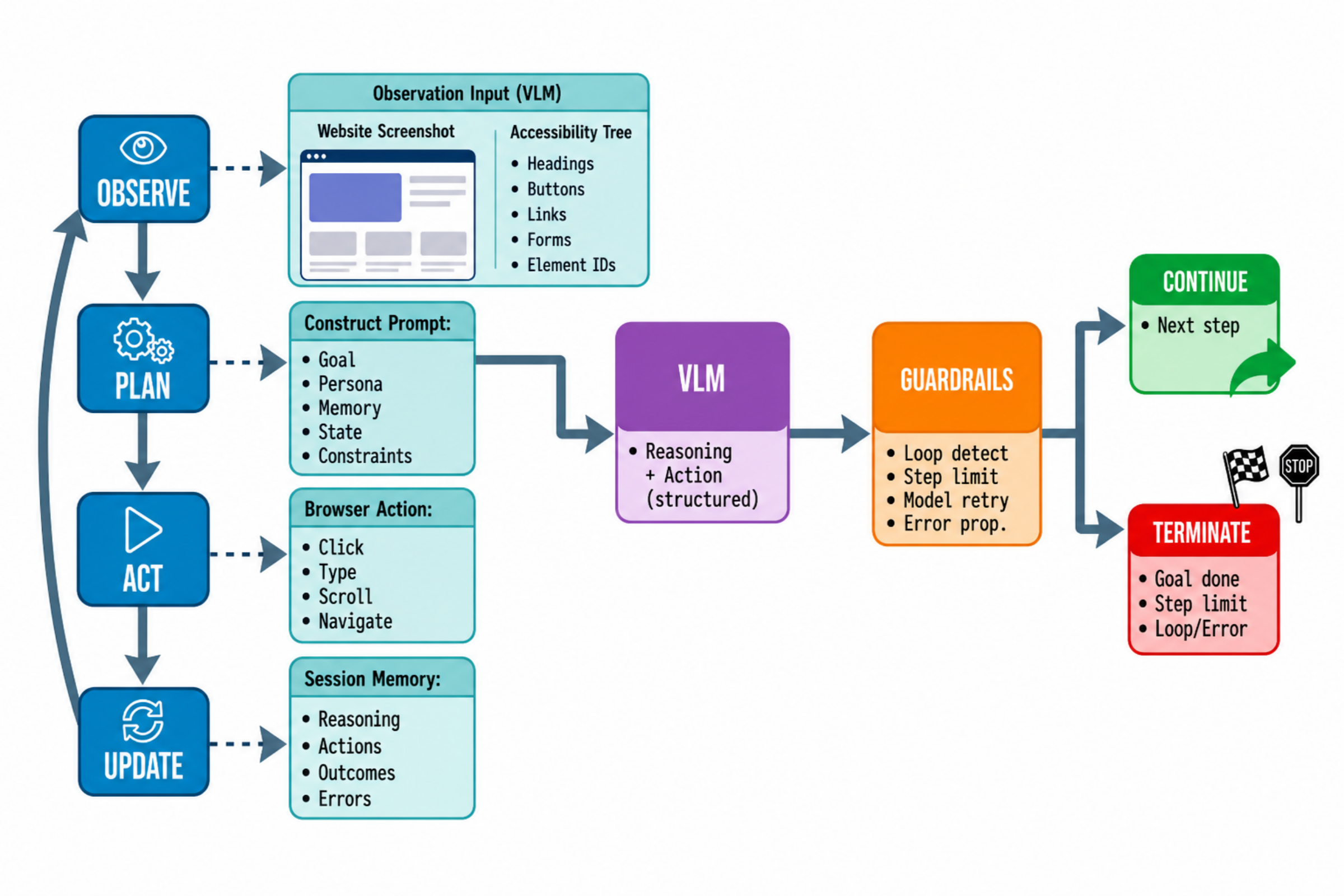}
    \caption{Agent Architecture.}
    \label{fig:agent_arch}
\end{figure}

\section{Agent Architecture}
\label{appendix:agent-arch}
\Cref{fig:agent_arch} visualizes the observe--plan--act loop introduced in
\Cref{sec: agent_arch}. At each step:
\begin{itemize}[nosep]
    \item \textbf{Observe.} The agent reads the current browser state
    (URL, accessibility tree, and screenshot) and retrieves the running
    session memory.
    \item \textbf{Plan.} The VLM produces a schema-constrained response
    containing its reasoning, a termination decision, and--if the session
    should continue--the next action to take.
    \item \textbf{Act.} The browser controller executes the action on the
    live storefront and writes the resulting page transition and outcome
    back to session memory for the next step.
\end{itemize}
An episodic session memory carries the full trajectory forward across steps
for temporal coherence, and a guardrail layer (loop protection, step and time
budgets, model retry, and error propagation) wraps the loop to enforce
graceful termination on failure. A full worked trajectory under this loop
is given in \Cref{sec: agent_trace}.

\section{Agent Trace Example} \label{sec: agent_trace}
\begin{figure}[t]
  \begin{center}
    \centerline{\includegraphics[width=0.5\columnwidth]{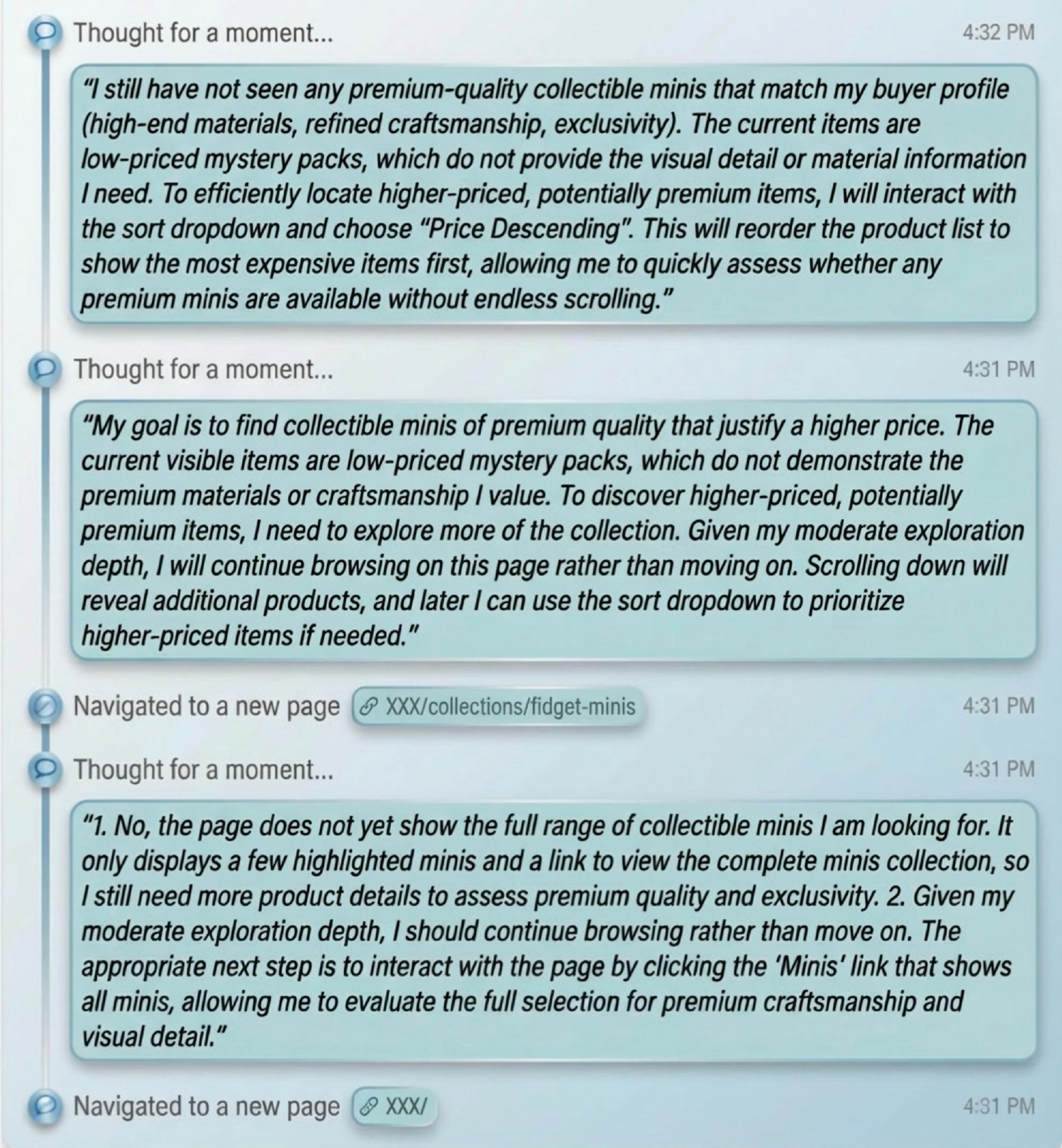}}
    \caption{Agent Reasoning During Initial Navigation.}
    \label{fig:trace_navigation}
  \end{center}
  \vskip -0.1in
\end{figure}

\begin{figure}[t]
  \vskip 0.1in
  \begin{center}
    \begin{subfigure}[b]{0.7\columnwidth}
      \centering
      \includegraphics[width=\textwidth, trim=8 20 20 20, clip]{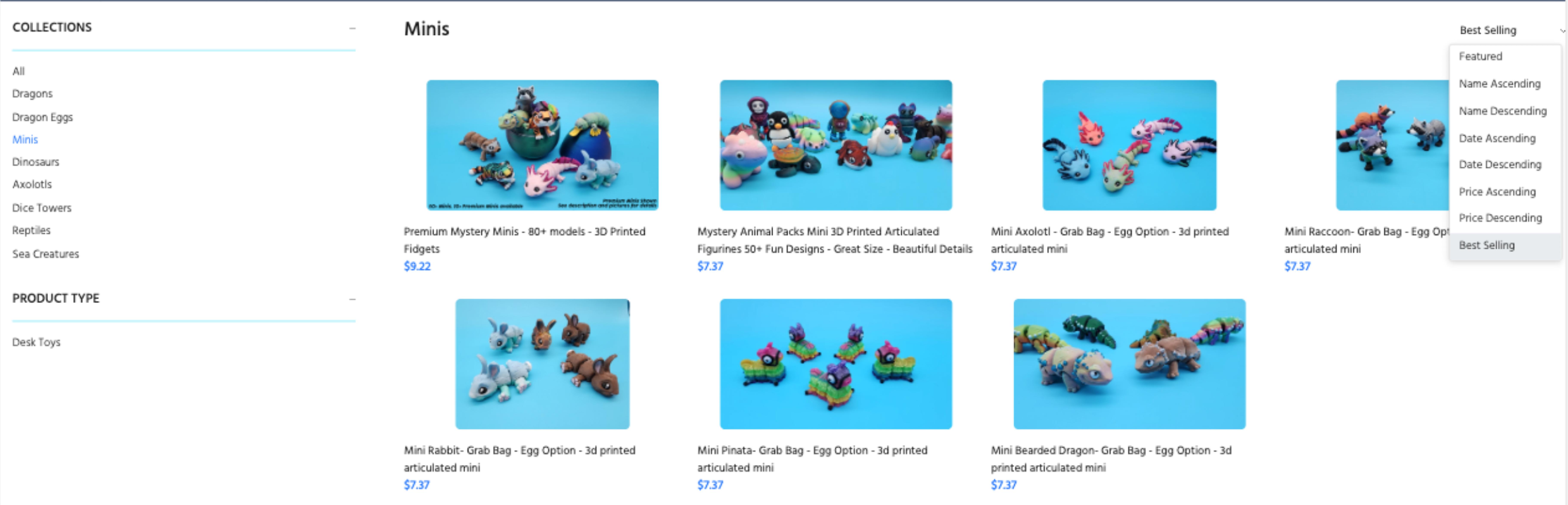}
      \caption{}
      \label{fig:minis_collection}
    \end{subfigure}
    \vspace{0.5em}
    \begin{subfigure}[b]{0.7\columnwidth}
          \centering
          \includegraphics[width=\textwidth, trim=8 20 20 20, clip]{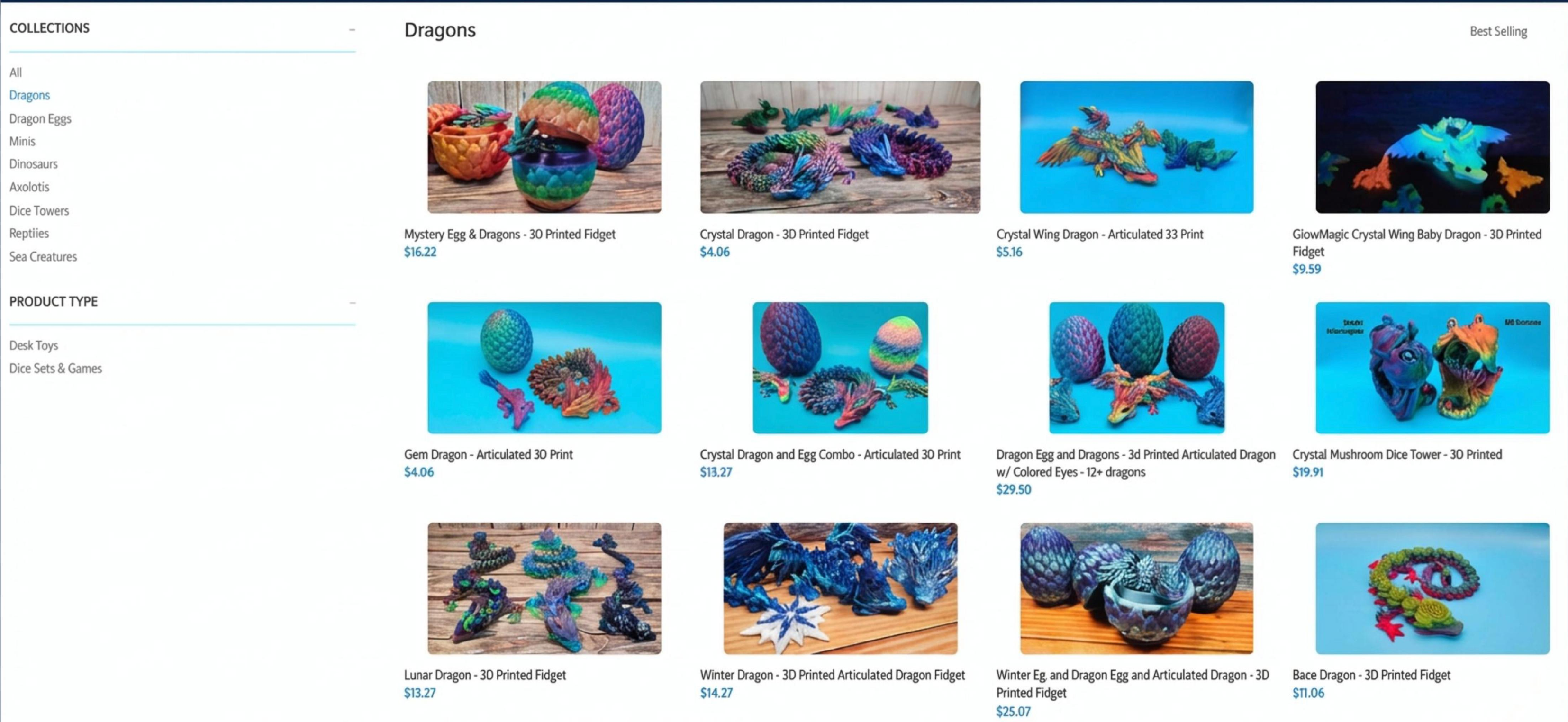}
          \caption{}
          \label{fig:dragons_collection}
    \end{subfigure}
    \vspace{0.5em}
    \begin{subfigure}[b]{0.7\columnwidth}
      \centering
      \includegraphics[width=\textwidth, trim=0 20 20 20, clip]{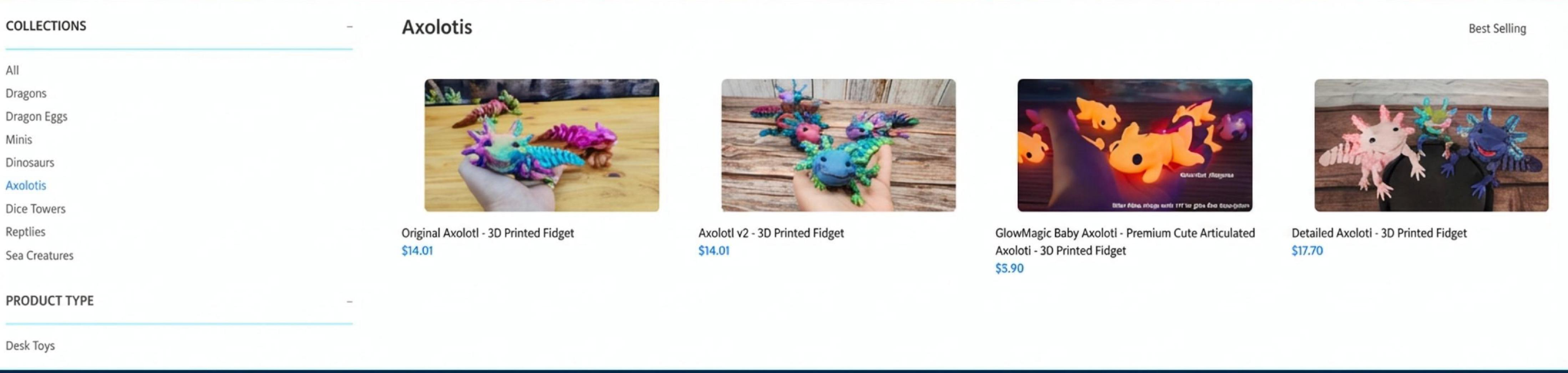}
      \caption{}
      \label{fig:axolotls_collection}
    \end{subfigure}
    \caption{Screenshots of Agent Browsing. (a) Minis collection (\$7–9). (b) Dragons collection (\$4–\$30). (c) Axolotls collection (\$6–\$18).}
    \label{fig:collections}
  \end{center}
  \vskip -0.1in
\end{figure}

\begin{figure}[t]
  \vskip 0.1in
  \begin{center}
    \centerline{\includegraphics[width=0.5\columnwidth]{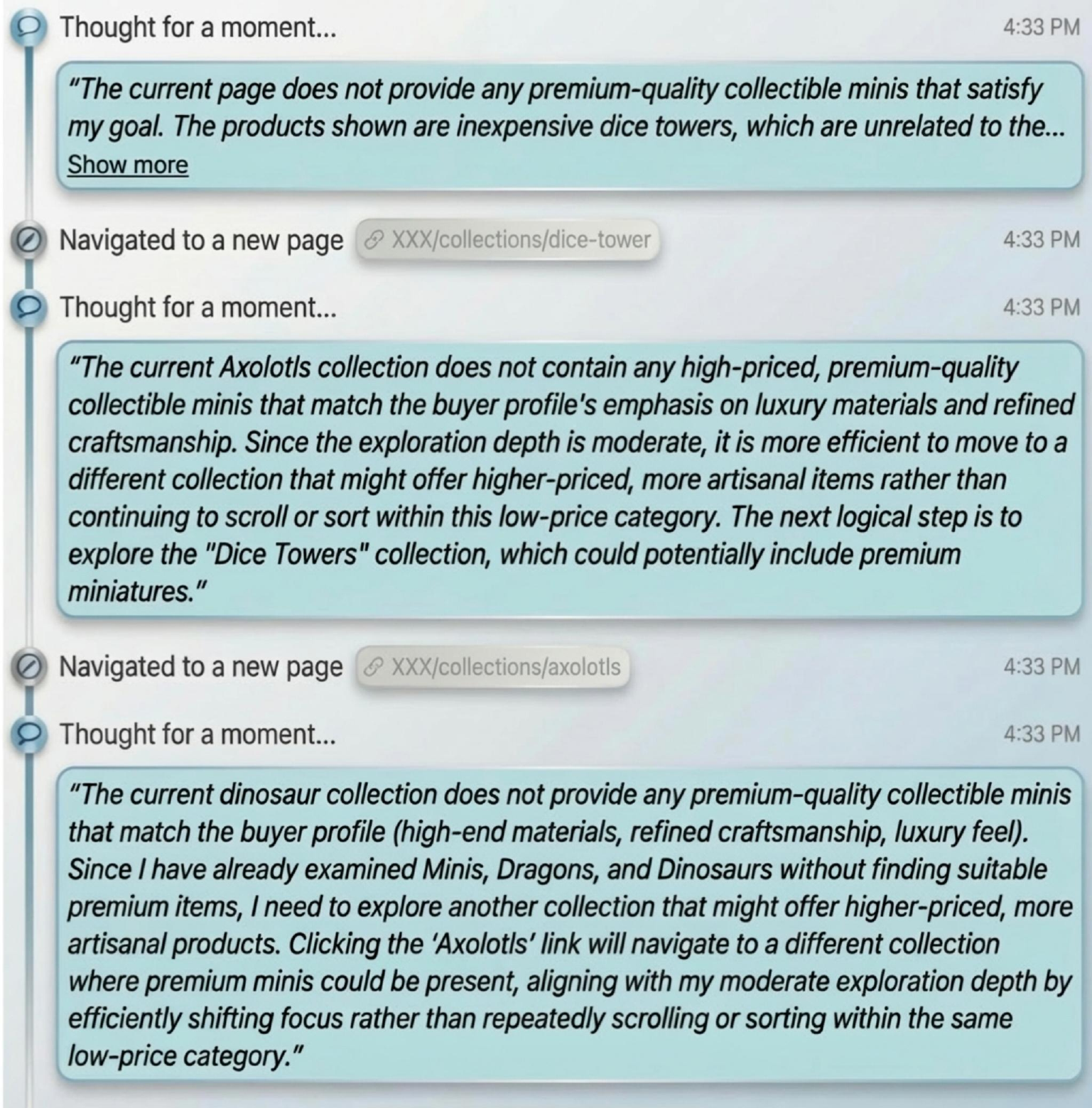}}
    \caption{Agent Reasoning During Collection Exploration.}
    \label{fig:trace_exploration}
  \end{center}
  \vskip -0.2in
\end{figure}

In this section, we present a complete trace of a \simgym~agent completing a shopping task on an anonymous store, specializing in $3D$ printed fidget toys and collectible figures. The agent's goal is to find and purchase a premium-quality collectible mini, with a buyer persona emphasizing luxury materials, refined craftsmanship, and moderate exploration depth.

\Cref{fig:trace_navigation} shows the agent's initial reasoning as it begins navigating the store. The agent explores multiple collections (\Cref{fig:collections}), systematically rejecting low-priced items that don't match its premium preferences. \Cref{fig:trace_exploration} captures the agent's reasoning as it moves between collections: \textit{``The current collection does not contain any high-priced, premium-quality collectible minis that match the buyer profile's emphasis on luxury materials and refined craftsmanship.''} Rather than settling for budget options priced at \$4–7, the agent efficiently shifts between collections, a behavior consistent with its moderate exploration depth.

\begin{figure}[t]
  \vskip 0.1in
  \begin{center}
    \begin{subfigure}[b]{0.5\columnwidth}
      \centering
      \includegraphics[width=\textwidth, trim=8 0 0 0, clip]{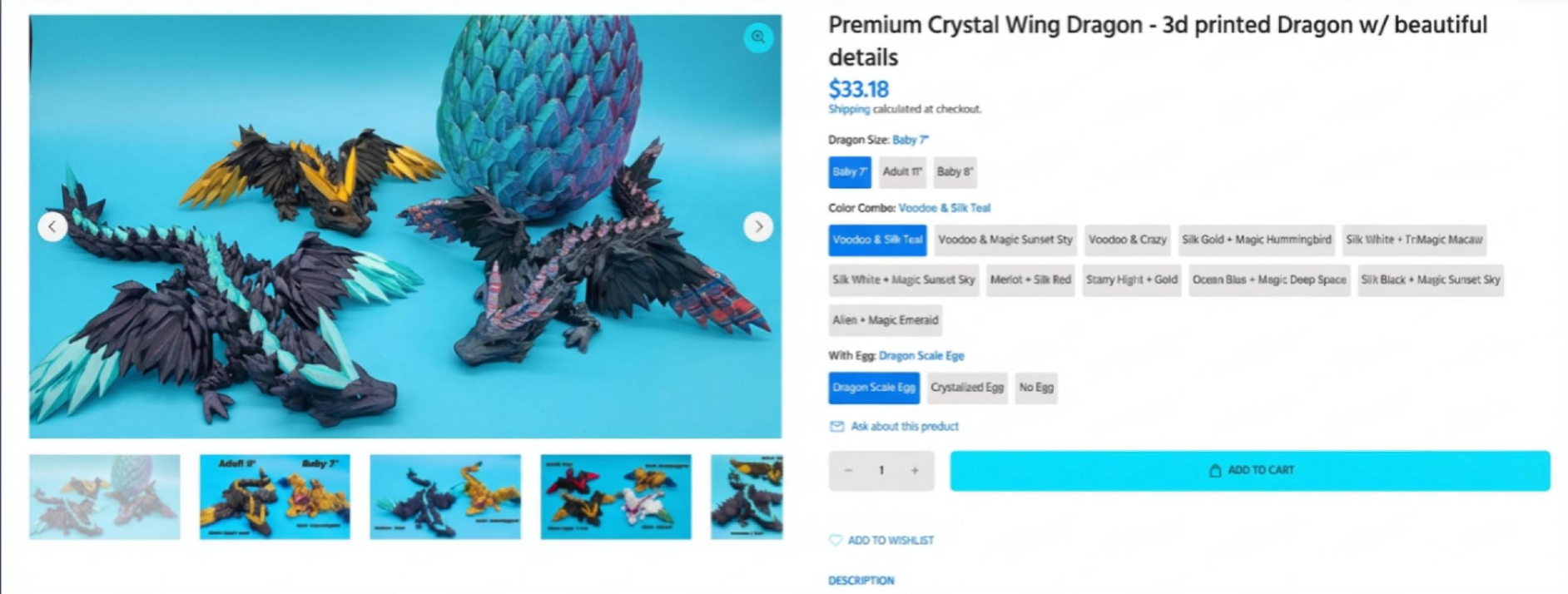}
      \caption{}
      \label{fig:product_page}
    \end{subfigure}
    \vspace{0.5em}
    \begin{subfigure}[b]{0.5\columnwidth}
      \centering
      \includegraphics[width=\textwidth]{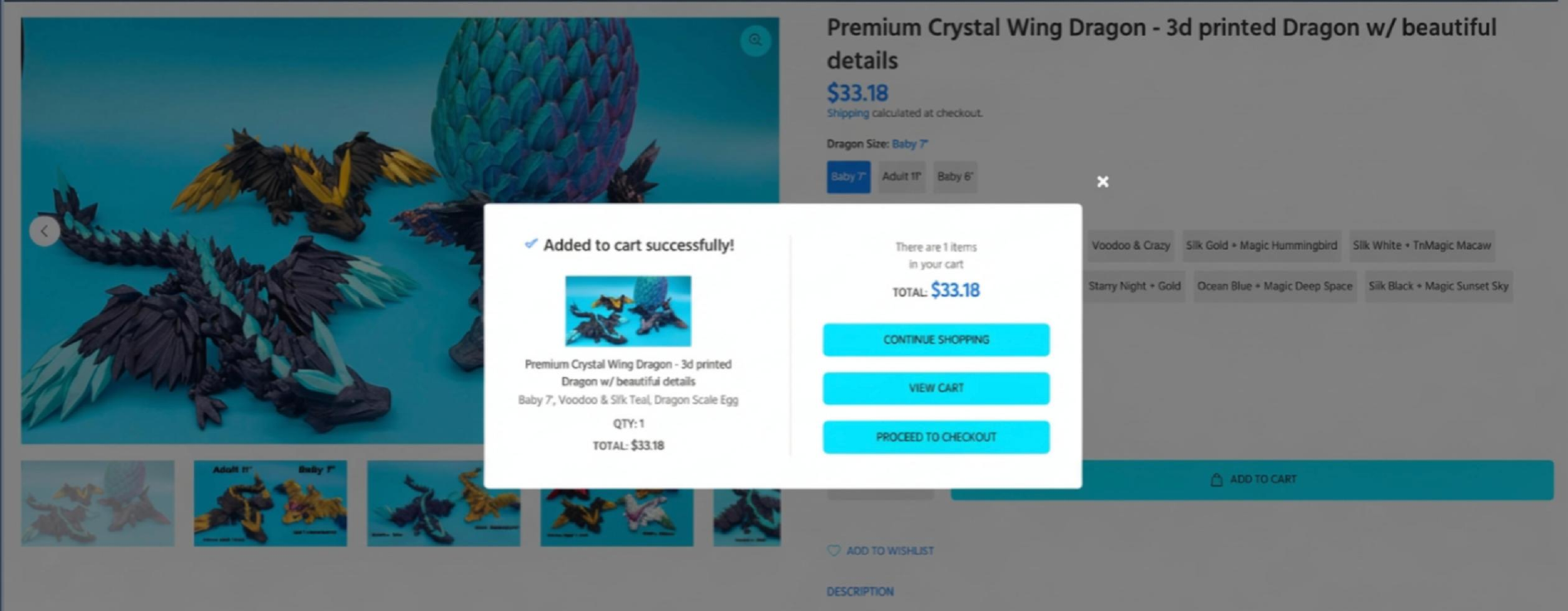}
      \caption{}
      \label{fig:cart}
    \end{subfigure}
    \caption{Product Selection and Add-to-cart. (a) Premium Crystal Wing Dragon product page (\$33.18). (b) Add-to-cart confirmation.}
    \label{fig:product_cart}
  \end{center}
  \vskip -0.1in
\end{figure}

\begin{figure}[t]
  \vskip 0.1in
  \begin{center}
    \centerline{\includegraphics[width=0.5\columnwidth]{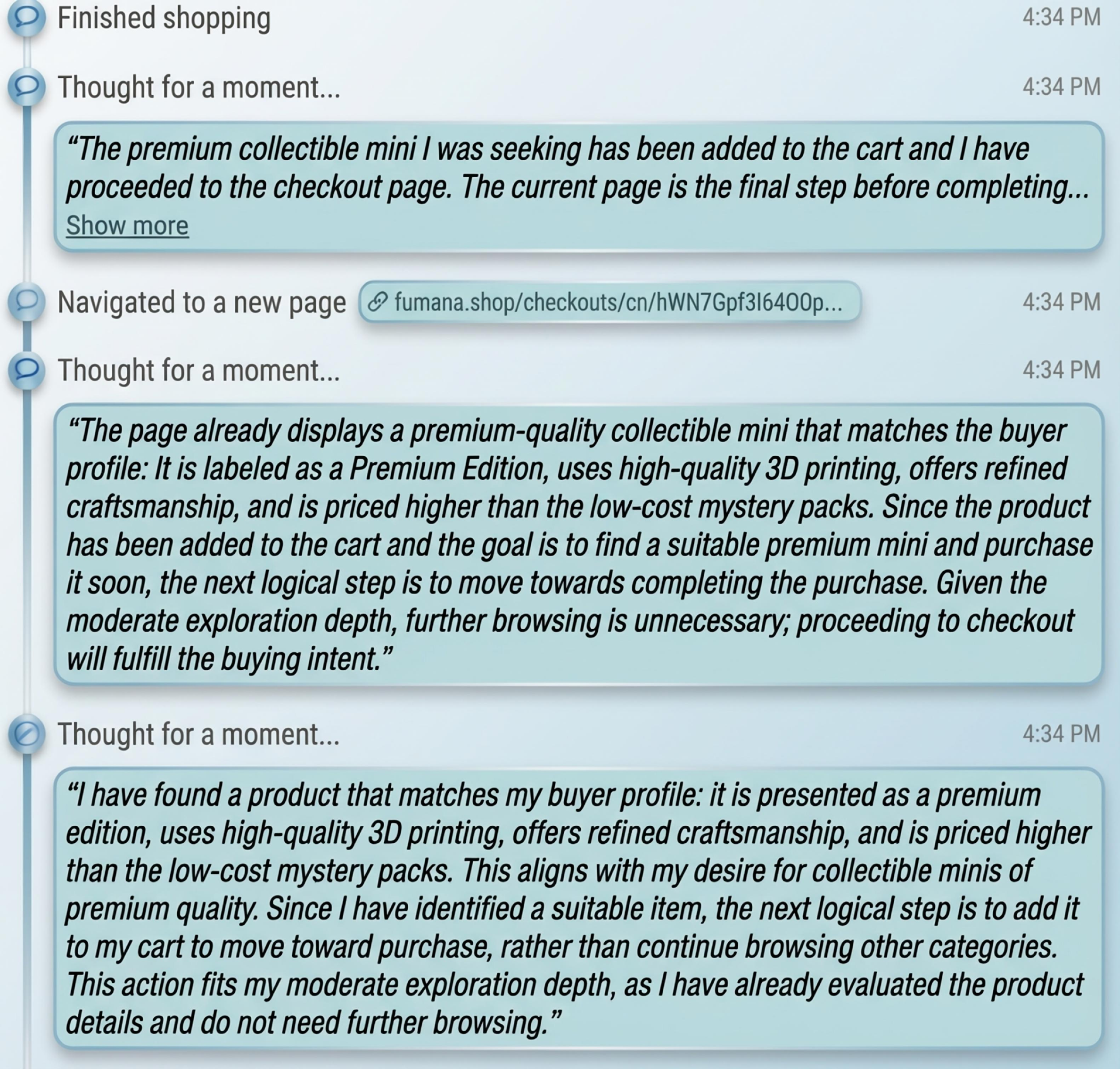}}
    \caption{Agent Reasoning During Purchase Decision and Checkout.}
    \label{fig:trace_checkout}
  \end{center}
  \vskip -0.1in
\end{figure}

After exploring multiple collections, the agent identifies a suitable product: the Premium Crystal Wing Dragon (\Cref{fig:product_page}), priced at \$33.18. The agent's reasoning confirms the match: \textit{``I have found a product that matches my buyer profile: it is presented as a premium edition, uses high-quality 3D printing, offers refined craftsmanship, and is priced higher than the low-cost mystery packs.''} The agent selects product options and adds the item to cart (\Cref{fig:cart}). \Cref{fig:trace_checkout} shows the final reasoning as the agent proceeds to checkout, concluding: \textit{``Given the moderate exploration depth, further browsing is unnecessary; proceeding to checkout will fulfill the buying intent.''}

This trace illustrates persona-consistent behavior throughout an extended shopping journey. The agent's premium preference led it to reject multiple budget-priced collections before identifying a suitable \$33.18 product, while its moderate exploration depth prevented excessive browsing once a match was found.

\begin{table}[t]
  \caption{Session-cluster distribution on the $50$-shop golden set. Skimmers (highlighted) is the cohort used throughout \Cref{sec:experiments-results}.}
  \label{tab:cohort-coverage}
  \begin{center}
    \begin{small}
        \begin{tabular}{lrrr}
          \toprule
          Cluster                    & \% of sessions & A2C rate & Bounce rate \\
          \midrule
          Immediate bouncers         & 59.1\%         & 0.0\%    & 100\%       \\
          \textbf{Skimmers (ours)}   & \textbf{26.4\%}& \textbf{9.5\%}  & 0\%   \\
          Engaged browsers           & 11.2\%         & 32.3\%   & 0\%         \\
          Purchase-ready             & 2.4\%          & 90.1\%   & 0\%         \\
          Deep researchers           & 0.9\%          & 27.8\%   & 0\%         \\
          \bottomrule
        \end{tabular}
    \end{small}
  \end{center}
  \vskip -0.1in
\end{table}

\begin{table}[t]
  \caption{Per-shop human $\Delta$A2C in the skimmers cohort vs.\ whole-shop signals (Pearson correlation across $50$ shops).}
  \label{tab:cohort-correlations}
  \begin{center}
    \begin{small}
        \begin{tabular}{lc}
          \toprule
          Comparison                                                & Pearson $r$ \\
          \midrule
          Skimmers vs.\ whole shop (all sessions)                   & 0.75 \\
          Skimmers vs.\ whole shop excl.\ immediate bouncers        & \textbf{0.89} \\
          \bottomrule
        \end{tabular}
    \end{small}
  \end{center}
\end{table}

\section{Cohort Coverage and Representativeness}
\label{appendix:cohort-coverage}

\Cref{tab:cohort-coverage} reports the cluster distribution of all sessions on the $50$-shop golden set, computed from the same session clustering used in \Cref{sec: intent_persona_pipeline}. The skimmers cohort is the largest cluster of \emph{engaged} (non-bouncing) sessions; the immediate-bounce cluster is excluded from any reported metric since it carries no UI-engagement signal, and the remaining three clusters are individually insignificant to support stable per-shop $\Delta$A2C estimates.

To check that reporting on a single cohort does not distort the shop-level picture, we compare the per-shop human $\Delta$A2C in the skimmers cohort against (i) the whole-shop signal across all sessions and (ii) the whole-shop signal with the immediate-bounce cluster excluded. Pearson correlation across the $50$ shops is $r{\approx}0.75$ for (i) and $r{\approx}0.89$ for (ii); see \Cref{tab:cohort-correlations}. The cohort signal is therefore a faithful proxy for the engaged shop-level signal.

\section{Confidence Interval Computation}
\label{appendix:ci}

We report $95\%$ confidence intervals for the two headline predictive-validity metrics, alignment rate and Pearson correlation, over the $n{=}50$ shops in the golden set. Each shop is the unit of analysis and contributes one paired $(\Delta\text{A2C}_{\text{human}}, \Delta\text{A2C}_{\text{sim}})$ observation, averaged across the two trials per shop (\Cref{sec:experiments-results}).

\textbf{Alignment rate.} We use the percentile bootstrap \cite{efron1979bootstrap} over shops. For each of $B{=}10{,}000$ resamples, we draw $n{=}50$ shops with replacement from the golden set and recompute alignment rate as the fraction of resampled shops on which the trial-averaged sign of the agent A2C shift agrees with the sign of the human A2C shift. The reported $95\%$ CI is the empirical $2.5$th--$97.5$th percentile interval over the $B$ bootstrap replicates.

\textbf{Pearson correlation.} We use the Fisher $z$-transform on the $n{=}50$ paired shop observations: $z = \tfrac{1}{2}\ln\!\left(\tfrac{1+r}{1-r}\right)$ with standard error $1/\sqrt{n-3}$, after which the symmetric $z$-interval is back-transformed to the $r$ scale. This is the standard parametric CI for the Pearson coefficient.

Both procedures share the same unit of analysis (the shop) and the same sample size ($n{=}50$), so the intervals are directly comparable across the two metrics within each table.


\end{document}